\documentclass[sigconf]{acmart}
\AtBeginDocument{%
  \providecommand\BibTeX{{%
    \normalfont B\kern-0.5em{\scshape i\kern-0.25em b}\kern-0.8em\TeX}}}

\usepackage{epsfig,graphicx,hyperref,amsfonts,xcolor,multirow,amsmath,balance}
\usepackage[misc]{ifsym} 

\acmConference[Conference acronym 'XX]{Make sure to enter the correct
  conference title from your rights confirmation emai}{June 03--05,
  2018}{Woodstock, NY}
\acmSubmissionID{2052}

\copyrightyear{2022} 
\acmYear{2022} 
\setcopyright{acmcopyright}\acmConference[MM '22]{Proceedings of the 30th ACM International Conference on Multimedia}{October 10--14, 2022}{Lisboa, Portugal}
\acmBooktitle{Proceedings of the 30th ACM International Conference on Multimedia (MM '22), October 10--14, 2022, Lisboa, Portugal}
\acmPrice{15.00}
\acmDOI{10.1145/3503161.3548246}
\acmISBN{978-1-4503-9203-7/22/10}

\begin{document}

\title{Improving Transferability for Domain Adaptive Detection Transformers}


\author{Kaixiong Gong}
\affiliation{%
  \institution{Beijing Institute of Technology}
  \state{Beijing}
  \country{China}
}
\email{kxgong@bit.edu.cn}

\author{Shuang Li$^{*}$}
\affiliation{%
  \institution{Beijing Institute of Technology}
  \state{Beijing}
  \country{China}
}
\email{shuangli@bit.edu.cn}

\author{Shugang Li}
\affiliation{%
  \institution{Beijing Institute of Technology}
  \state{Beijing}
  \country{China}
}
\email{shugangli@bit.edu.cn}

\author{Rui Zhang}
\affiliation{%
  \institution{Beijing Institute of Technology}
  \state{Beijing}
  \country{China}
}
\email{zhangrui20@bit.edu.cn}

\author{Chi Harold Liu}
\affiliation{%
  \institution{Beijing Institute of Technology}
  \state{Beijing}
  \country{China}
}
\email{liuchi02@gmail.com}

\author{Qiang Chen}
\affiliation{%
  \institution{Baidu VIS}
  \state{Beijing}
  \country{China}
}
\email{chenqiang13@baidu.com}

\thanks{$^{*}$Shuang Li is the corresponding author.}

\renewcommand{\shortauthors}{Kaixiong Gong et al.}

\begin{abstract}
  
  DETR-style detectors stand out amongst in-domain scenarios, but their properties in domain shift settings are under-explored. This paper aims to build a simple but effective baseline with a DETR-style detector on domain shift settings based on two findings. For one, mitigating the domain shift on the backbone and the decoder output features excels in getting favorable results. For another, advanced domain alignment methods in both parts further enhance the performance. Thus, we propose the Object-Aware Alignment (OAA) module and the Optimal Transport based Alignment (OTA) module to achieve comprehensive domain alignment on the outputs of the backbone and the detector. The OAA module aligns the foreground regions identified by pseudo-labels in the backbone outputs, leading to domain-invariant base features. The OTA module utilizes sliced Wasserstein distance to maximize the retention of location information while minimizing the domain gap in the decoder outputs. We implement the findings and the alignment modules into our adaptation method, and it benchmarks the DETR-style detector on the domain shift settings. Experiments on various domain adaptive scenarios validate the effectiveness of our method.

\end{abstract}





\maketitle

\section{Introduction}

\begin{figure}[t]
    \centerline{\includegraphics[width=0.95\linewidth]{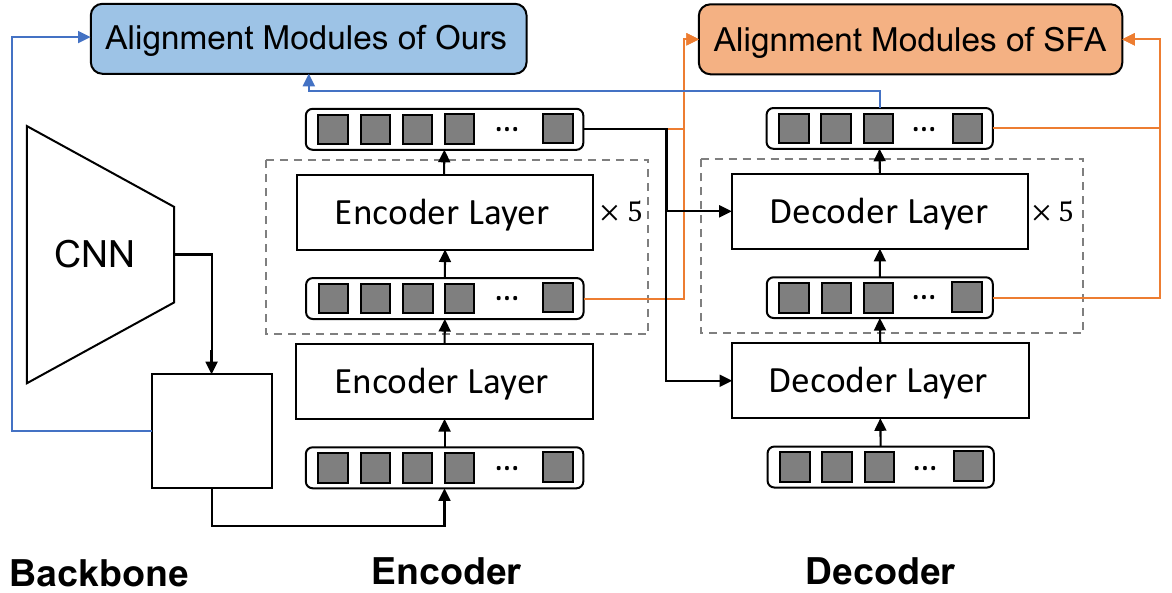}}
    \caption{Overview of the alignment paradigms of ours and SFA \cite{wang2021exploring}. Ours aligns the output features of CNN backbone and decoder. While SFA aligns the token embeddings of each layer of encoder and decoder.}
    \label{fig:fig_oaa_sfa}
    \vspace{-6mm}
\end{figure}

\begin{figure*}[t]
    \centerline{\includegraphics[width=0.9\linewidth]{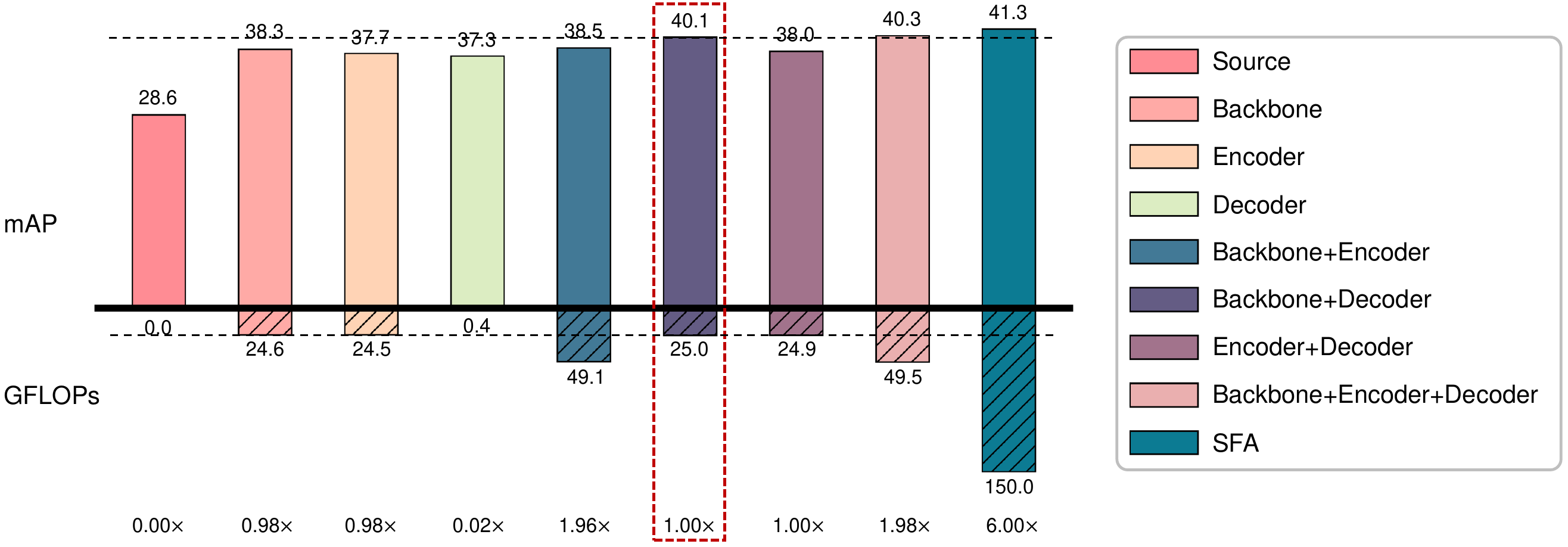}} \vspace{-4mm}
    \caption{The top shows the mAPs of different alignment variations on Cityscapes $\rightarrow$ Foggy Cityscapes, while the bottom presents the extra computational cost of these variations by excluding the overhead of the based detector. ``Source'' denotes the basic Deformable DETR model \cite{zhu2020deformable}. ``Backbone'', ``Encoder'' and ``Decoder'' denote aligning the multi-scale features of the CNN backbone, the output features of encoder and decoder via adversarial training \cite{dann}, respectively. The rest variations denote unifying any two or all alignment modules for training. This work adopts the ``Backbone+Decoder'' scheme (highlighted with the red dash) at the consideration of performance and overhead.}
    \label{fig:fig_intro}
\end{figure*}

\begin{figure}[t]
    \centerline{\includegraphics[width=0.9\linewidth]{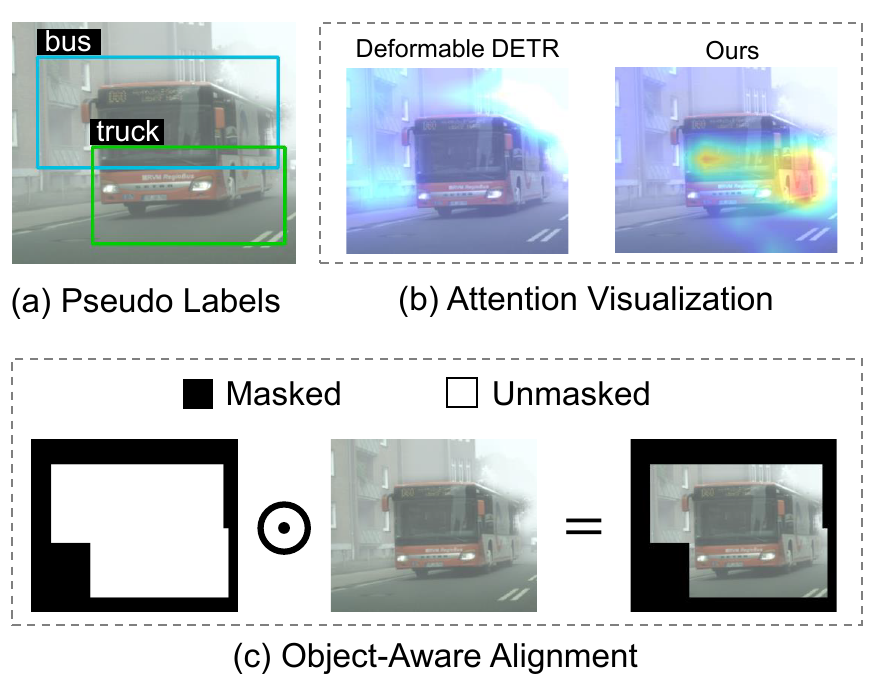}}
    \caption{(a): Cases of inaccurate pseudo labels, misclassification (green box) and mislocation (green and blue boxes). (b): Attention visualizations for target samples of source model and our method. Ours facilitates the CNN backbone concentrating on the foreground regions via Object-Aware Alignment, while the attention of the source model is biased to the background area. (c): Illustration of Object-Aware Alignment (OAA). The pixels of the feature map enclosed by the pseudo bounding boxes will be highlighted during alignment. We use the pseudo labels in a class-agnostic manner. Unifying two inaccurate pseudo boxes is still able to uncover the object region. Consequently, OAA reduces the negative effect of inaccurate pseudo labels and make full use of them.}
    \label{fig:fig_oaa}
    \vspace{-4mm}
\end{figure}

DETR-style detectors \cite{detr,zhu2020deformable} remove hand-craft designs in the standard in-domain scenarios and offer promising performance. 
However, the domain shift scenarios with changing weather, varying illumination, and altering scene layout are more related to reality. In this paper, we aim to extend the application scenarios of DETR-style detectors by creating a simple but effective baseline for domain adaptive object detection (DAOD).
In DAOD, methods and recipes for DETR-style detectors are yet to be built, compared with the well-established  CNN-based detectors. That brings us back to the basics and asks the question: \textit{where and how to adapt with DETR-style detectors?}

Wang et al. \cite{wang2021exploring} did their practice, advocating aligning the sequence token embeddings of each layer of encoder and decoder, demonstrating the potential of detection transformers in the domain adaptive setting, as shown in Fig.\ref{fig:fig_oaa_sfa}. Nonetheless, we argue that it's cumbersome and sub-optimal to align the embeddings of each layer of encoder and decoder. 

The first concern is about ``where''. The detection transformer comprises three main networks: a CNN backbone to extract the feature representations from raw images, an encoder for feature enhancement, and a decoder to probe features for detecting objects. 
In this work, we propose to align the output features of the backbone and the decoder, diminishing the domain gap at both pixel and instance levels. The reasons are: 1) The CNN backbone processes images and outputs the features that will be served as the input of the subsequent transformer. Consequently, the quality of the CNN features is the cornerstone of detection performance. 2) Encoder enhances the backbone features by aggregating pixels of feature maps. It still generates pixel level of features but in a sequence style so that the decoder can better conduct cross-attention for detecting objects. In addition, DETR model has the issue of slow convergence due to the limitation of the attention module on processing image feature maps. An extra feature alignment objective might hardly coordinate with the feature enhancement during optimization. Hence, it's not essential to align the encoder feature with extra computational overhead. 3) The output features of the decoder are directly utilized for regression and classification, which are crucial for the final performance. However, the decoder features are inevitably biased to the source domain due to the absence of target supervision. Hence, we follow \cite{yosinski2014transferable}\footnote{It suggests that the features of deeper layers are more transferable.} and propose to align the output features of it instead of the token embeddings of each layer.

We empirically support this claim by conducting comparison experiments of various alignment scheme variations, as shown in Fig. \ref{fig:fig_intro}. Only performing alignment on one network obtains unsatisfactory performance gains. Taking a step further, aligning the output features of the backbone and the decoder leads to favorable performance. Furthermore, one can observe that backbone alignment is essential for reducing the domain gap, while encoder alignment provides limited improvement.

The second concern is about ``how''. SFA \cite{wang2021exploring} exploits adversarial learning for aligning the token embeddings of each decoder layer. Although effective, adversarial alignment might distort the intrinsic distribution characteristics \cite{bsp}. In the case of decoder alignment, this adversarial scheme might damage the crucial location information, which limits the performance of detectors.

\label{section:introduction}

To fully boost the cross-domain performance, we propose an Object-Aware Alignment (OAA) module for acquiring domain-invariant backbone features. Conducting global alignment on backbone feature maps effectively narrow the domain shift. Still, it overlooks the difference between the objects and the background. The object regions should be emphasized during alignment since they are more transferable than the background regions in which the domain shift is usually presented. In light of this, we introduce a simple yet effective scheme, utilizing the pseudo labels to identify the foreground regions, which is complementary to the global alignment. Concretely, the pixels of feature maps enclosed by pseudo bounding boxes will be highlighted. As a result, the transformer will focus more on the object area of target samples, as illustrated in Fig. \ref{fig:fig_oaa}(b), leading to a higher recall. Prior works generally leverage the pseudo labels for self-training \cite{kim2019self,jiang2021decoupled}, which might suffer from performance degeneration due to the noisy pseudo labels, as illustrated in Fig. \ref{fig:fig_oaa}(a). However, in OAA, we utilize the pseudo labels in a class-agnostic manner (see Fig. \ref{fig:fig_oaa}(c)). Hence, the noisy pseudo labels can still be helpful in identifying the desired regions that contain objects, resulting in in-depth backbone feature alignment.

On the other hand, the decoder features are also severely biased to the source domain due to the absence of supervision from target labels. The features of the decoder carry important location information, which is crucial for final predictions. Hence, how to reduce the domain gap while simultaneously maintaining the location characteristics is critical for domain adaptive detection transformers. Instead of applying adversarial learning, we resort to Wasserstein Distance, which takes into account the intrinsic geometry information. Concretely, Wasserstein Distance seeks an optimal transport plan for moving one distribution to another with minimal cost. Thus, adopting it to mitigate the discrepancy between source and target domains on the decoder features might be able to maximumly preserve the location information. Nonetheless, the complexity of approximating Wasserstein Distance in the high-dimensional feature space is intolerable. To overcome this, we utilize sliced Wasserstein Distance, solving several one-dimensional optimal transport problems by projecting the decoder features to one-dimensional space, which have closed-form solutions. Thus, the detector can be efficiently trained, and the location information is reserved. Consequently, our method yields more highly accurate bounding boxes, as verified in Section \ref{mislocation}. Extensive experiments on several adaptive benchmarks manifest that our method consistently enhances the detection performance on target domains of the DETR-style detector and surpasses various competitive approaches.

In summary, the main contributions of this work are:
\begin{enumerate}
    \item We discovered that mitigating domain shift on the output features of CNN backbone and decoder yields favorable results with cost-efficient computational overhead.
  
    \item We propose an Object-Aware Alignment (OAA) and an Optimal Transport based Alignment (OTA) modules for reducing the domain shift at pixel and instance levels.
    \item Extensive experiments on several adaptation benchmarks validate the effectiveness of our method.
\end{enumerate}

\section{Related Work}
In this section, we discuss the prior works related to ours. 

\textbf{Object detection}.
Object detection is a fundamental task in the computer vision field. In the early years, CNN-based detectors dominated the object detection community, which can be roughly divided into two categories: two-stages \cite{faster-rcnn}, and one-stage detectors \cite{yolo,ssd}. Two-stages detectors first generate proposals with selective search \cite{fast_rcnn} or region proposal network \cite{faster-rcnn}, then produce predictions by refining the proposals. While one-stage detectors directly predict detection results from images, thus enjoying faster inference speed.

Recently, a transformer-based detector, DETR \cite{detr} was introduced to achieve fully end-to-end detection and eliminate the hand-craft designed components, such as anchor generation and non-maximum suppression, which attracted a surge of research interest. Following, Deformable DETR \cite{zhu2020deformable} develops a sparse attention module named deformable attention to fasten the convergence speed of DETR. Sharing the same spirit, many researchers \cite{up-detr,dn_detr,dino_detr,dab_detr} proposed various schemes to speed up the convergence of DETR. More recently, Wang et al. \cite{wang2022towards} pointed out that DETR has the issue of data hunger and proposed to solve it by augmenting the supervision. Nonetheless, how to enhance the generalization ability of the DETR-style models to new domains has not been fully explored yet. In this work, we delve into enhancing the transferability of DETR-style models.

\textbf{Domain adaptive object detection.}
DAF \cite{dafaster} is the first work attempting to transfer the object detector across domains, which proposes to align the backbone and the ROI features. Inspired by it, researchers are devoted to boost the transferability of detectors by improving feature alignment schemes \cite{strong-weak,scda,gpadet,xu2020exploring,maf,divmatch,proda,ViSGA,I3Net,RPA}. SWDA \cite{strong-weak} proposes to employ strong alignment on the local features and weak alignment on the global features of the CNN backbone. Zhang et al. \cite{RPA} introduce learnable RPN prototypes for aligning the RPN features. MTOR \cite{mtor} and UMT \cite{deng2021unbiased} exploit the teacher-student framework to overcome domain shift. Wang et al. \cite{wang2022robust} further adapt the detection model towards target domains with open and compound degradation styles. While above works are designed for Faster RCNN \cite{faster-rcnn}, some other approaches are built on one-stage detectors \cite{kim2019self,hsu2020every}. 

With the rising of DETR-style detectors \cite{detr,zhu2020deformable}, Wang et al. \cite{wang2021exploring} advocate aligning the token embeddings of each layer of encoder and decoder to improve the cross-domain performance of them. 
By contrast, we find that aligning the output features of the CNN backbone and decoder yields favorable results and leads to cost-efficient computational overhead.

\textbf{Pseudo labels.}
Adopting pseudo labels for training is frequent in domain adaptation classification. Typically, the algorithms leverage the confident target prediction for calibrating the parameters of classifiers, leading to better conditional distribution alignment \cite{CBST,MSTN}. Further, \cite{SHOT,liang2021domain} refine the pseudo labels via clustering the target data. This is infeasible in detection transformers since the decoder features contain important location information with large intra-class variance. In the context of DAOD, some prior works \cite{kim2019self,jiang2021decoupled} utilize the pseudo bounding boxes and classification results for self-training, which heavily relies on the quality of pseudo labels. By contrast, we adopt pseudo labels to mine the foreground regions to emphasize the importance of objects during alignment. Furthermore, our method is more robust for noisy pseudo labels, as illustrated in Fig. \ref{fig:fig_oaa}.

\textbf{Optimal transport.} 
Optimal transport has been applied in domain adaptation (DA) for transforming the source distribution to the target one \cite{flamary2016optimal,jdot,deepjtot}. Prior works \cite{flamary2016optimal,jdot,deepjtot} aim to reach an optimal transport plan of aligning the distributions of two domains. Derived from optimal transport theory, Wasserstein Distance (WD), is a metric that measures the minimal cost of the optimal transport problem, which has been used in generative adversarial nets \cite{WGAN} and domain adaptation \cite{shen2018wasserstein}. Since it's non-trivial to directly approximate the WD in high-dimensional space, sliced Wasserstein Distance (SWD) \cite{Wasserstein} is then proposed to approximate WD in the one-dimensional space via projecting the original vectors to one-dimensional space, which enjoys the geometry property of Wasserstein Distance and efficient estimation. Chen et al. \cite{lee2019sliced} utilize SWD to enhance the cross-domain performance of the classifier under the DA setting. Different from them focusing on image recognition, we concentrate on the more difficult task, object detection. Ge et al. \cite{ge2021ota} leverage optimal transport to improve the label assignment of object detectors. By contrast, we propose two alignment modules for mitigating the domain gap on both pixel and instance levels of features. To the best of our acknowledge, we are the first to introduce SWD to reduce the domain gap in DAOD.

\section{The Proposed Method}
In this section, we present our proposed method. We start by introducing the problem setting and the based detector. In the context of domain adaptive object detection (DAOD), we can access a set of labeled source data $D_s=\{ \boldsymbol{x}_{i}^s, Y_i^s \}_{i=1}^{n_s}$, where $\boldsymbol{x}_i^s$ is the source image, $ Y_i^s = \{(b_i^1, c_i^1), (b_i^2, c_i^2), ..., (b_i^{m_i}, c_i^{m_i})\}$ is the annotation set containing $m_i$ bounding boxes $b_i$ and corresponding categories $c_i$. Meanwhile, the target domain only includes fully unlabeled target data $D_t=\{\boldsymbol{x}_i^t\}_{i=1}^{n_t}$. Since the weather, illumination and layout vary across domains, there exists crucial domain shift that impairs the detection performance on target domain. The goal of DAOD is to enhance the generalization performance of detectors on the target domain using labeled source data and unlabeled target data.

In this work, we build our method on the Deformable DETR \cite{zhu2020deformable}, which consists of a CNN backbone network for extracting based features, an encoder for feature enhancement, a decoder for feature probing, and finally feed-forward network (FFN) for prediction. In order to better detect small objects, Deformable DETR exploits the multi-scale features of the CNN backbone, where the small objects are detected from the high-resolution features. Let $\{f^l\}_{l=1}^{L}$\footnote{L is set as 4. The feature maps of stages $C_3$ to $C_5$ of backbone plus one extra feature map generated by applying a convolution on stage $C_5$ are adopted.} denotes the $l^{th}$ level of feature, where $f^l \in \mathbb{R}^{C^l \times W^l \times H^l} $. In the following, the multi-scale backbone features will be flattened, embedded, and enhanced with positional and level embeddings \cite{zhu2020deformable} to construct sequence input for the transformer. Encoder further refines the features using the attention mechanism. Learnable object queries are utilized to probe the encoder features for detecting objects in the decoder. Finally, FFN predicts the bounding boxes and categories of objects based on the decoder features.

As discussed in Section \ref{section:introduction}, the quality of backbone features is the cornerstone of detection performance on target domains. On the other hand, the output features of the decoder are directly utilized for detection. However, unfortunately, the features of the decoder are inevitably skewed to the source domain due to the absence of supervision from the target domain. To remedy these issues, we propose two alignment modules: an Object-Aware Alignment (OAA) module for CNN backbone features and an Optimal Transport based Alignment (OTA) module for decoder features.

\subsection{Object-Aware Alignment}
Domain-invariant backbone features are crucial for detection transformers, which will ease the domain shift issue. In Deformable DETR, the multi-scale backbone features are applied to improve the detection performance for small objects. Likewise, in order to better detect the small objects of the target domain, we establish feature alignment on multi-scale backbone features. The features of different level $\{f^l\}_{l=1}^{L}$ are fed into a domain discriminator, generating the domain scores for each pixel.
 If we denote the $P^l=\{p^l_i \in \mathbb{R}^{W^l \times H^l} |i=1,2,...,n_s+n_t \}$ as the discriminator outputs for $l^{th}$ level of backbone features,  we can formulate the adversarial training objective on $l^{th}$ level features of backbones as:

\begin{align}
    \mathcal{L}_{d}^{l} = \mathop{\sum}_{i,u,v} \log (p_i^{s(u,v)}) + ( 1-  \log (p_i^{t(u,v)}) ), p_i^s,p_i^t \in P^l,
    \label{eq:backbone-l-level}
\end{align}where the $p_i^{s(u,v)}, p_i^{t(u,v)}$ denote the discriminator outputs for source and target images located at $(u,v)$, respectively. Based on Eq. \eqref{eq:backbone-l-level}, we could define the alignment objective on multi-scale features:
\begin{align}
    \mathcal{L}_{d} = \mathop{\sum}_{l=1}^{L} \mathcal{L}_d^l.
    \label{eq:backbone}
\end{align}

To align the backbone feature distributions, the domain discriminator is optimized by minimizing Eq. \eqref{eq:backbone}, while the backbone is updated for maximizing it. This global alignment strategy is beneficial to mitigating the domain gap across domains. However, in the context of detection, this alignment scheme is unsatisfactory since 1) the foreground regions should be emphasized to reduce false negative detections and 2) the foreground regions are more transferable than the background ones in which domain gap is usually presented. Thus, we further utilize the predictions of FFN to highlight the foreground regions that contain objects since no region proposals are available in detection transformers. In other words, we encourage the detector to concentrate on the local regions enclosed by the predicted boxes. Former works typically adopt the pseudo labels for self-training, which might lead to error accumulation due to noise labels. As we adopt the pseudo labels for feature alignment in a class-agnostic manner, the negative effect of the errors could be mitigated, as shown in Fig. \ref{fig:fig_oaa}.

Technically, let $ \hat{Y}_i^t = \{(\hat{b}_i^1, \hat{s}_i^1), (\hat{b}_i^2, \hat{s}_i^2), ..., (\hat{b}_i^{m_i}, \hat{s}_i^{m_i})\}$ denote the predicted results for $i^{th}$ target image, where $\hat{s}_i^j$ is the confidence score for $j^{th}$ detection result, i.e., the maximum prediction probability over object classes. Note that we abuse the notation $m_i$ here for simplicity. For acquiring confident detections, a threshold $\tau$ is utilized to filter the pseudo labels. Consequently, we obtain a pseudo bounding box set $\hat{B}^t_i = \{\hat{b}_i^j | \hat{s}_i^j  > \tau, (\hat{b}_i^j, \hat{s}_i^j) \in \hat{Y}_i^t \}$ for $i^{th}$ target image, which is exploited as the indicator for re-weighting the importance of each pixel of target backbone features:
$$ w^{t(u,v)}_i = \left\{
\begin{aligned}
  1&, if \ (u,v) \ located \ within \ \hat{B}^t_i, \\
  0&, else.
\end{aligned}
\right.
$$
The weights are utilized to encourage the backbone network to focus on those foreground areas and learn the patterns of objects across domains.
Likewise, we directly exploit the ground truth bounding boxes of source samples to generate weights for source samples.
Unifying the weights with alignment loss, we fulfill object-aware alignment: 

\begin{align}
    \hat{\mathcal{L}}_{d}^{l} &= \mathop{\sum}_{i,u,v} w^{s(u,v)}_{i}  \log (p_i^{s(u,v)}) \\ \nonumber
    &+ w^{t(u,v)}_{i} ( 1-  \log (p_i^{t(u,v)}) ), p_i^s,p_i^t \in P^l. 
\end{align}

Via aggregating $\hat{\mathcal{L}}_{d}^{l}$ on each level of features, we acquire $\hat{\mathcal{L}}_{d}$ for performing object-aware alignment on multi-scale features. On the basis of global alignment loss, we build the object-aware alignment loss. The two losses are used for adapting backbone features, as they are complementary to each other. Global alignment narrows the gap for better pseudo labels, while the region alignment improves detection for foreground objects. The loss of OAA module is:
\begin{align}
    \mathcal{L}_{OAA} = \mathcal{L}_d + \lambda \hat{\mathcal{L}}_d,
\end{align}where $\lambda$ is a trade-off hyper-parameter to balance the contributions of the two losses.

\subsection{Optimal Transport Based Alignment}
The features produced by decoder are directly utilized for predicting the location and category, which are critical for the detection performance. Unfortunately, the decoder is inevitably biased towards the source domain. Importantly, adversarial alignment across domains might impair the intra-class variance \cite{bsp}, resulting in weak representation ability for the location information.

To reduce the domain shift and simultaneously preserve the location information, we resort to Wasserstein Distance, a metric derived from optimal transport theory, calculating the minimal cost of transporting a distribution to another. Wasserstein Distance seeks an optimal transport plan and leverages it to measure the cost. Formally, if we denote two random variables $z_1, z_2 \in \mathbb{R}^d$, a set of all joint distributions $\gamma (z_1,z_2)$ as $\varGamma (\mu,\nu )$. $\gamma (z_1,z_2)$ indicates that how much ``mass'' is transported from $z_1$ to $z_2$ for converting distribution $\mu$ to $\nu$. The Wasserstein Distance can be formulated as:
\begin{align}
    W(\mu,\nu) = \mathop{\inf}_{\gamma \in \varGamma(\mu,\nu)} \int_{\gamma} c(z_1, z_2)d(z_1,z_2),
\end{align}where $c(\cdot,\cdot)$ is the cost function, such as Squared Euclidean distance. 

However, it's non-trivial to estimate the optimal transport plan $\gamma^{*}$ in the high-dimensional space, i.e., the feature space. To overcome this, sliced Wasserstein Distance (SWD) is utilized \cite{Wasserstein} for approximating the Wasserstein Distance, which solves optimal transport problems in the one-dimensional space via projecting the features to it. In specific, we sample $K$\footnote{We set $K=256$.} projection vectors from a unit sphere $\mathbb{S}=\{\theta \in \mathbb{R}^d| \| \theta\|=1 \}$, where $d$ is the dimension of decoder features. Then, these projection vectors will be exploited to project the decoder features to one-dimensional space. Let $f_{dec}=\{f_{dec}^1, f_{dec}^2, ..., f_{dec}^N\}$ denote the decoder features, the SWD of source and target domains is defined as:
\begin{align}
    \mathcal{L}_{OTA} = \mathop{\sum}_{k=1}^{K} \mathop{\sum}_{i}^{N} \| sort(\theta^{\top}_k f_{dec}^{s(i)}) - sort(\theta^{\top}_k f_{dec}^{t(i)})   \|_2^2,
    \label{eq:ota}
\end{align}where $sort(\cdot)$ is  the sorting function that ranks the elements from small to big values, $f_{dec}^{s(i)}, f_{dec}^{t(i)}$ are the $i^{th}$ decoder feature from source and target domains. The projection vector $\theta_k$ maps the decoder features into one-dimensional space, i.e., $\theta^{\top}_k f_{dec}^{s(i)}, \theta^{\top}_k f_{dec}^{t(i)} \in \mathbb{R}^1$. Hence, we in fact solve the one-dimensional optimal transport problems by optimizing Eq. \eqref{eq:ota}, which have closed-form solutions. Consequently, the domain gap on decoder features will be effectively reduced and the retention for location information is achieved. As validated in \ref{mislocation}, OTA facilitates the detector to generate highly accurate bounding boxes.

\subsection{Total Loss}
Here, we introduce the total loss involved in training our method. First, the supervised detection loss is employed on source labeled data for learning the source knowledge. Second, the adaptation losses, i.e., the losses of OAA and OTA modules, are applied to facilitate the knowledge transfer from source to target domains. The overall loss of ours:
\begin{align}
    \mathcal{L} = \mathcal{L}_{det} + \mathcal{L}_{OAA} + \beta \mathcal{L}_{OTA}
\end{align}where $\mathcal{L}_{det}$ is the supervised detection loss on source domain, $\beta$ is the trade-off hyper-parameter.

\begin{table}[t]
            \centering
            \small
            \caption{Results on synthetic to real adaptation scenario, i.e., Sim10k $\rightarrow$ Cityscapes. D-DETR denotes Deformable DETR \cite{zhu2020deformable}.} 
            \label{tab:sim10k}
            \setlength{\tabcolsep}{4mm}
            \begin{tabular}{c|c|c} 
                \toprule[1.0pt]
                Method & Detector & \emph{car} AP \\
                
                \hline
                Faster RCNN (source) & Faster RCNN & 34.6 \\
                DAF \cite{dafaster} & Faster RCNN & 41.9 \\
                DivMatch \cite{divmatch} & Faster RCNN & 43.9 \\
                SWDA \cite{strong-weak} & Faster RCNN & 44.6 \\
                SCDA \cite{scda} & Faster RCNN & 45.1 \\
                MTOR \cite{mtor} & Faster RCNN & 46.6 \\
                CR-DA~\cite{xu2020exploring} & Faster RCNN & 43.1 \\
                RPA~\cite{RPA} & Faster RCNN & 45.7 \\
                CR-SW~\cite{xu2020exploring} & Faster RCNN & 46.2 \\
                GPA  \cite{gpadet} & Faster RCNN & 47.6 \\
                ViSGA  \cite{ViSGA} & Faster RCNN & 49.3 \\
                D-adapt  \cite{jiang2021decoupled} & Faster RCNN & 53.2 \\
                \hline
                FCOS \cite{tian2019fcos} (source) & FCOS & 42.5 \\
                EPM~\cite{hsu2020every} & FCOS & 47.3 \\
                KTNet~\cite{KTNet} & FCOS & 50.7 \\
                \hline
                Deformable DETR (source) & D-DETR & 47.4  \\
                SFA \cite{wang2021exploring} & D-DETR & 52.6 \\
                O$^2$net (ours) & D-DETR & \textbf{54.1} \\
                \bottomrule[1.0pt]
            \end{tabular}
        
\end{table}

\section{Experiments}
In this section, we conduct extensive experiments to evaluate our method. We name our approach as O$^2$net as it contains OAA and OTA which are validated by an ablation study and visualization analysis. Code will be available at \url{https://github.com/BIT-DA/O2net}.

\begin{table*}[t]
            \centering
            \small
            \caption{Results on weather adaption scenario , i.e., Cityscapes $\rightarrow$ Foggy Cityscapes. D-DETR denotes Deformable DETR \cite{zhu2020deformable}.} 
            \label{tab:foggy}
            \setlength{\tabcolsep}{3.2mm}
            \begin{tabular}{c|c|cccccccc|c}
                \toprule[1.0pt]
                Method & Detector & person & rider & car & truck & bus & train & mcycle & bicycle & mAP \\
            
                \hline
                Faster RCNN (source) & Faster RCNN & 26.9 & 38.2 & 35.6 & 18.3 & 32.4 & 9.6 & 25.8 & 28.6 & 26.9 \\
                DAF \cite{dafaster} & Faster RCNN & 29.2 & 40.4 & 43.4 & 19.7 & 38.3 & 28.5 & 23.7 & 32.7 & 32.0 \\
                DivMatch \cite{divmatch} & Faster RCNN & 31.8 & 40.5 & 51.0 & 20.9 & 41.8 & 34.3 & 26.6 & 32.4 & 34.9 \\
                SWDA \cite{strong-weak} & Faster RCNN & 31.8 & 44.3 & 48.9 & 21.0 & 43.8 & 28.0 & 28.9 & 35.8 & 35.3 \\
                SCDA \cite{scda} & Faster RCNN & 33.8 & 42.1 & 52.1 & 26.8 & 42.5 & 26.5 & 29.2 & 34.5 & 35.9 \\
                MTOR \cite{mtor} & Faster RCNN & 30.6 & 41.4 & 44.0 & 21.9 & 38.6 & 40.6 & 28.3 & 35.6 & 35.1 \\
                CR-DA~\cite{xu2020exploring} & Faster RCNN & 30.0 & 41.2 & 46.1 & 22.5 & 43.2 & 27.9 & 27.8 & 34.7 & 34.2 \\
                CR-SW~\cite{xu2020exploring} & Faster RCNN & 34.1 & 44.3 & 53.5 & 24.4 & 44.8 & 38.1 & 26.8 & 34.9 & 37.6 \\
                GPA \cite{gpadet} & Faster RCNN & 32.9 & 46.7 & 54.1 & 24.7 & 45.7 & 41.1 & 32.4 & 38.7 & 39.5 \\
                D-adapt \cite{jiang2021decoupled} & Faster RCNN & 40.8 & 47.1 & 57.5 & \textbf{33.5} & 46.9 & 41.4 & 33.6 & 43.0 & 43.0 \\
                ViSGA \cite{ViSGA} & Faster RCNN & 38.8 & 45.9 & 57.2 & 29.9 & \textbf{50.2} & \textbf{51.9} & 31.9 & 40.9 & 43.3 \\
                \hline
                FCOS \cite{tian2019fcos} (source) & FCOS & 36.9 & 36.3 & 44.1 & 18.6 & 29.3 & 8.4 & 20.3 & 31.9 & 28.2 \\
                EPM~\cite{hsu2020every} & FCOS & 44.2 & 46.6 & 58.5 & 24.8 & 45.2 & 29.1 & 28.6 & 34.6 & 39.0 \\
                KTNet~\cite{KTNet} & FCOS & 46.4 & 43.2 & 60.6 & 25.8 & 41.2 &40.4 & 30.7 & 38.8 & 40.9 \\
                \hline
                Deformable DETR (source) & D-DETR & 38.0 & 38.7 & 45.3 & 16.3 & 26.7 & 4.2 & 22.9 & 36.7  & 28.6 \\
                SFA \cite{wang2021exploring} & D-DETR & 46.5 & 48.6 & 62.6 & 25.1 & 46.2 & 29.4 & 28.3 & 44.0 & 41.3 \\
                O$^2$net (ours) & D-DETR & \textbf{48.7} & \textbf{51.5} & \textbf{63.6} & 31.1 & 47.6 & 47.8 & \textbf{38.0} & \textbf{45.9} & \textbf{46.8} \\
                \bottomrule[1.0pt]
            \end{tabular}
        \vspace{-2mm}
\end{table*}

\begin{table*}[t]
            \centering
            \small
            \caption{Results on scene adaptation scenario, i.e., Cityscapes $\rightarrow$ BDD100k. D-DETR denotes Deformable DETR \cite{zhu2020deformable}.}
            \label{tab:bdd}
            \setlength{\tabcolsep}{3.2mm}
            \begin{tabular}{c|c|ccccccc|c}
                \toprule[1.0pt]
                Method & Detector & person & rider & car & truck & bus & mcycle & bicycle & mAP \\
            
                \hline
                Faster R-CNN (source) & Faster RCNN & 28.8 & 25.4 & 44.1 & 17.9 & 16.1 & 13.9 & 22.4 & 24.1 \\
                DAF~\cite{dafaster} & Faster RCNN & 28.9 & 27.4 & 44.2 & 19.1 & 18.0 & 14.2 & 22.4 & 24.9 \\
                SWDA~\cite{strong-weak} & Faster RCNN & 29.5 & 29.9 & 44.8 & 20.2 & 20.7 & 15.2 & 23.1 & 26.2 \\
                SCDA~\cite{scda} & Faster RCNN & 29.3 & 29.2 & 44.4 & 20.3 & 19.6 & 14.8 & 23.2 & 25.8 \\
                CR-DA~\cite{xu2020exploring} & Faster RCNN & 30.8 & 29.0 & 44.8 & 20.5 & 19.8 & 14.1 & 22.8 & 26.0 \\
                CR-SW~\cite{xu2020exploring} & Faster RCNN & 32.8 & 29.3 & 45.8 & \textbf{22.7} & 20.6 & 14.9 & \textbf{25.5} & 27.4 \\
                \hline
                FCOS~\cite{tian2019fcos} (source) & FCOS & 38.6 & 24.8 & 54.5 & 17.2 & 16.3 & 15.0 & 18.3 & 26.4 \\
                EPM~\cite{hsu2020every} & FCOS & 39.6 & 26.8 & 55.8 & 18.8 & 19.1 & 14.5 & 20.1 & 27.8 \\
                \hline
                Deformable DETR (source) & D-DETR & 38.9 & 26.7 & 55.2 & 15.7 & 19.7 & 10.8 & 16.2 & 26.2 \\
                SFA \cite{wang2021exploring} & D-DETR &  40.2 & 27.6 & 57.5 & 19.1 & 23.4 & \textbf{15.4} & 19.2 & 28.9 \\
                O$^2$net (ours) & D-DETR & \textbf{40.4} & \textbf{31.2} & \textbf{58.6} & 20.4 & \textbf{25.0} & 14.9 & 22.7 & \textbf{30.5} \\
                \bottomrule[1.0pt]
            \end{tabular}
        \vspace{-2mm}
    \end{table*}

\subsection{Experiment Setup}
The following datasets are adopted for evaluation:
\textbf{Cityscapes}. The Cityscapes \cite{cityscapes} is composed of city scene images. The training set of Cityscapes contains 2,975 images while the validation set has 500 images. We adopt the validation set to test methods when Cityscapes is the target domain.
\textbf{Foggy Cityscapes}. Foggy Cityscapes \cite{foggy} is the fog version of Cityscapes by applying fog synthesis algorithm on the original Cityscapes images for generating foggy images. Thus, Foggy Cityscapes and Cityscapes share identical annotations. 
\textbf{Sim10k}. Powered by the game engine of Grand Auto Theft, Sim10k \cite{sim10k} contains 10,000 generated images of the game scenes as well as 58,701 annotations for cars. The images of Sim10k can be leveraged to construct the synthetic to real adaption scenario.
\textbf{BDD100k}. BDD100k \cite{yu2018bdd100k} includes 70,000 training images and 10,000 validation images. The subset of BDD100k that contains \textit{daytime} images is selected to construct the adaption benchmark. Thus, we use 36,728 images for training and 5,258 images for validation.

With the aforementioned datasets, we build three adaptation scenarios: weather adaption (Cityscapes $\rightarrow$ Foggy Cityscapes), synthetic to real adaptation (sim10k $\rightarrow$ Cityscapes) and scene adaptation (Cityscapes $\rightarrow$ BDD100k). Following DAF \cite{faster-rcnn}, we report the results of mean Average Precision (mAP) with a threshold $0.5$.

\subsection{Implementation Details}
We adopt the Deformable DETR \cite{zhu2020deformable} with a ImageNet \cite{deng2009imagenet} pre-trained ResNet-50 \cite{resnet} backbone as the based detector. Inherited from the original paper \cite{zhu2020deformable}, the learning rate of Deformable DETR is set as 2 $\times$ $10^{-4}$ and the Adam optimizer \cite{kingma2014adam} is used for updating parameters. The learning rate of the domain discriminator adopted in OAA module is 4 $\times$ $10^{-3}$. As \cite[]{jiang2021decoupled}, we pre-train models on source domain for a reliable initialization for pseudo labels. The threshold $\tau$ is set as $0.5$ for all experiments. $\lambda$ and $\beta$ are both $1$ on Cityscapes $\rightarrow$ Foggy Cityscapes, and $0.1$ on the rest.

\begin{table*}[t]
    \centering
    \small
    \caption{Ablation study on Cityscapes $\rightarrow$ Foggy Cityscapes. OAA and OTA denote the proposed Object-Aware Alignment and Optimal Transport based Alignment modules. GA and ADA indicate global alignment on backbone features and adversarial alignment on decoder features.} \label{tab:ablation}
    \setlength{\tabcolsep}{2mm}
    \begin{tabular}{c|cccc|cccccccc|c}
        \toprule[1.0pt]
        Methods & OAA & OTA & GA & ADA  & person & rider & car & truck & bus & train & mcycle & bicycle & mAP \\
        \hline
        Deformable DETR (source) &  &  &  &  & 38.0 & 38.7 & 45.3 & 16.3 & 26.7 & 4.2 & 22.9 & 36.7  & 28.6 \\
        \hline
        \multirow{4}{*}{Proposed} & $\checkmark$ &  &  &  & 45.8 & 48.7 & 61.2 & 30.4 & 48.3 & 33.5 & 34.0 & 43.1 & 43.1 \\
         &  & $\checkmark$ &  &  & 46.2 & 47.5 & 61.9 & 24.1 & 45.9 & 29.2 & 25.1 & 41.8 & 40.2 \\
         &  & $\checkmark$ & $\checkmark$ &  & 47.4 & 49.3 & 63.8 & 28.9 & 46.5 & 32.6 & 32.1 & 44.7 & 43.2 \\
         & $\checkmark$ &  &  & $\checkmark$ & 48.3 & 50.5 & 65.0 & 30.7 & 49.5 & 34.5 & 34.6 & 44.3 & 44.7 \\
        \hline
        
        O$^2$net (ours) & $\checkmark$ & $\checkmark$ &  &  & 48.7 & 51.5 & 63.6 & 31.1 & 47.6 & 47.8 & 38.0 & 45.9 & 46.8 \\
        \bottomrule[1.0pt]
    \end{tabular}
\vspace{-4mm}
\end{table*}

\begin{figure*}[t]
    \centerline{\includegraphics[width=0.95\linewidth]{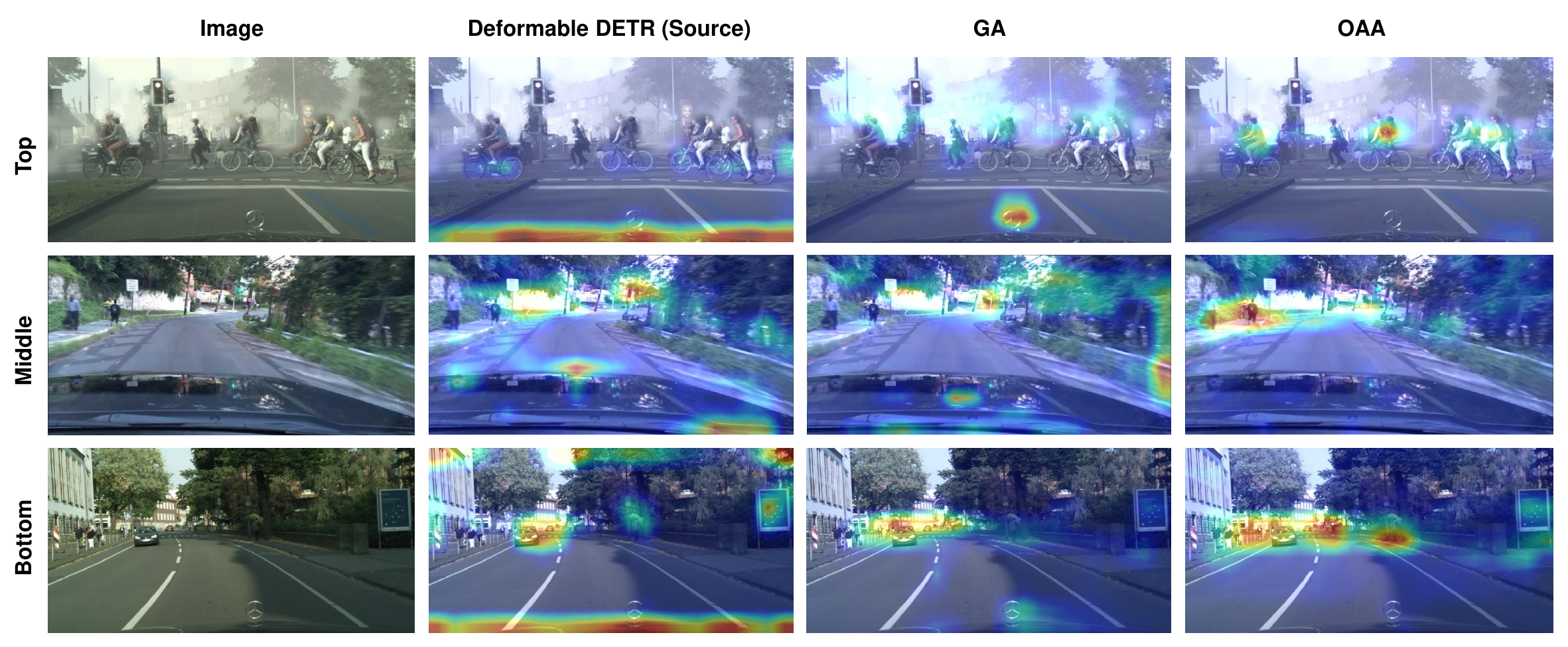}} \vspace{-4mm}
    \caption{Illustration of the attention maps for testing samples. We presented the attention maps of three methods, i.e., Deformable DETR (source), Global Alignment (GA) and Object-Aware Alignment (OAA) on backbone features. From top to bottom, we show the testing examples from Cityscapes $\rightarrow$ Foggy Cityscapes, Cityscapes $\rightarrow$BDD100k and Sim10k $\rightarrow$ Cityscapes.}
    \label{fig:fig_heatmap}
    \vspace{-4mm}
\end{figure*}

\subsection{Comparisons with SOTA Methods}

\textbf{Weather adaptation}. Handling varying weather is crucial and frequent in autonomous driving. Thus, we evaluate our method on Cityscapes $\rightarrow$ Foggy Cityscapes. Table \ref{tab:foggy} shows the results, from which one can observe that ours significantly exceeds others, validating that the two proposed alignment modules can indeed mitigate the domain shift under the variation of weather conditions.

\textbf{Synthetic to real adaptation}. Adopting synthetic images for training is economical, saving the cost of collecting and annotating data. Sim10k $\rightarrow$ Cityscapes reflects how well the model adapted from synthesis to real domains. Table \ref{tab:sim10k} shows that our method achieves promising performance improvement over other methods.

\textbf{Scene adaptation}. To test the robustness to the layout variation, we conduct the experiments on Cityscapes $\rightarrow$ BDD100k as shown in Table \ref{tab:bdd}. O$^2$net consistently improves the source model, validating the alignment strategies are general to scene adaption.

\begin{figure}[t]
    \centerline{\includegraphics[width=0.8\linewidth]{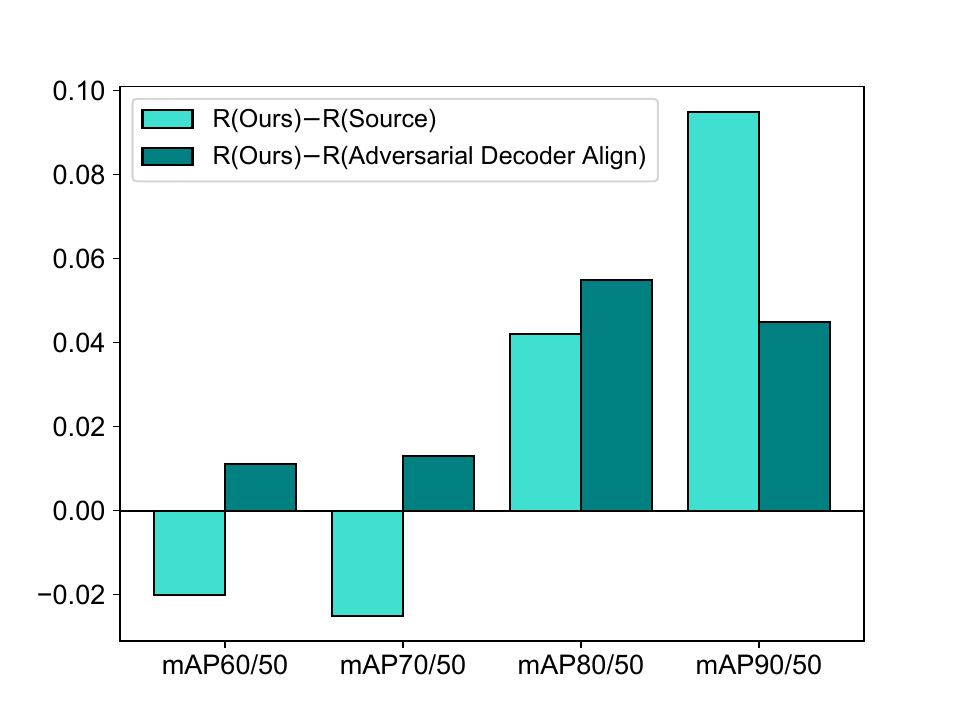}}
    \vspace{-2mm}
    \caption{The difference of the mAP ratio (R) that measures the percentage of highly accurate predicted bounding boxes in all accurate bounding boxes. For example, mAP70$/$50 indicates the ratio of predicted bounding boxes with IOU $\geq 0.7$ in the predicted bounding boxes with IOU $\geq 0.5$. ``R(Ours)$-$R(Source)'' denote the map ratios of ours minus the map ratios of the source model.}
    \label{fig:fig_mislocation}
    \vspace{-4mm}
\end{figure}

\begin{figure*}[t]
    \centerline{\includegraphics[width=0.95\linewidth]{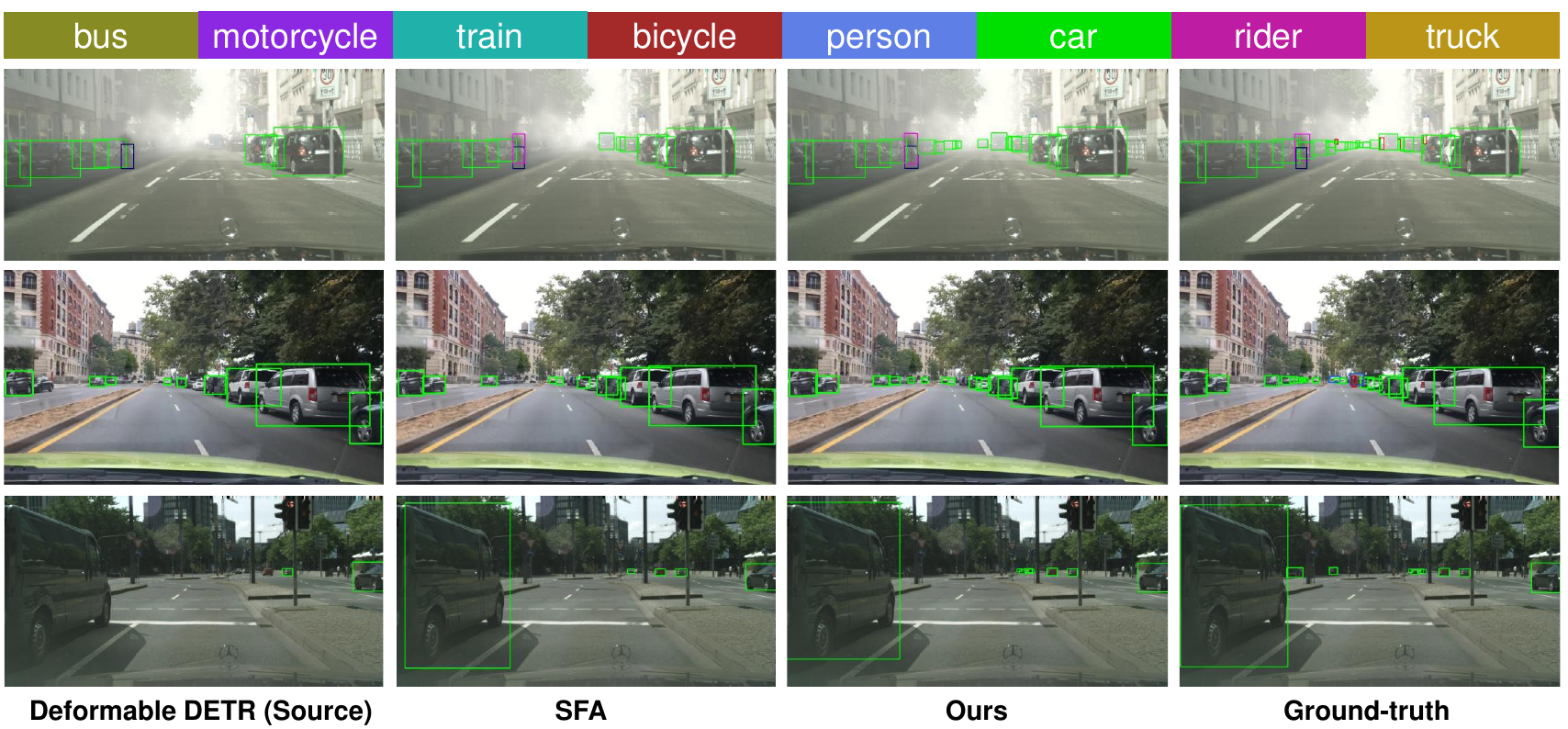}}
    \caption{Qualitative results: Top, middle and bottom exhibit the visualization detection results of adaptation scenario Cityscapes $\rightarrow$ Foggy Cityscapes, Cityscapes $\rightarrow$ BDD100k and Sim10k $\rightarrow$ Cityscapes.}
    \label{fig:fig_qualitative}
    \vspace{-4mm}
\end{figure*}

\begin{figure}[t]
    \centerline{\includegraphics[width=0.95\linewidth]{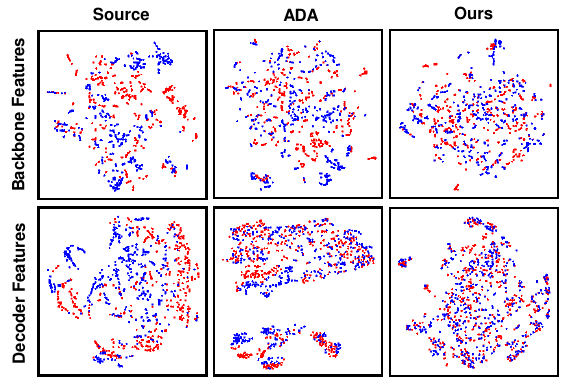}}
    
    \caption{t-SNE \cite{tsne} visualization of the features of CNN backbone and decoder. The red and blue dotes denote the features from source and target domains, respectively. Adversarial decoder alignment (ADA) indicates that utilizing adversarial training to align the features of decoder.}
    \label{fig:fig_tsne}
    \vspace{-6mm}
\end{figure}

\subsection{Ablation Study} 
To validate each component of our method, an ablation study is conducted on Cityscapes $\rightarrow$ Foggy Cityscapes (see Table \ref{tab:ablation}). We build our O$^2$net by adding components to the source model. First, when adding OAA, the detection performance significantly improves, manifesting the importance of domain-invariant backbone feature maps. Second, adding OTA, the performance gain is less impressive than OAA. The interpretation is that the biased backbone features severely hinder the adaptation of the following transformer. Via adding both the alignment modules, we reach our proposed method, which yields the best performance. One could observe that the two alignment modules are complementary to each other. OAA provides domain-invariant features at pixel level while the OTA further reduces the domain gap at instance level.

In addition, we analyze the two modules by introducing their variations, i.e., global alignment (GA) on backbone and adversarial decoder alignment (ADA). The former enhances the model performance but will be more powerful when concentrating on the foreground regions. On the other hand, the latter achieves inferior results compared with OTA, revealing that ADA might distort the location information, resulting in limited performance.

\subsection{Analytical Experiments}

\textbf{Attention visualization}. 
The attentions for backbone feature maps extracted by the source model, global alignment (GA) and our OAA are in Fig. \ref{fig:fig_heatmap}. GA improves the source model via globally aligning the backbone features. By contrast, OAA produces more high attentions for objects, thus boosting the true positives. Though the backbone also puts some attention on the background, it will be refined by the transformer. Meanwhile, OAA further reduces the attention for background area while emphasizing as more objects as possible, which leads to higher precision.

\textbf{Analysis of highly accurate bounding boxes}.
\label{mislocation}
SWD is utilized to narrow the domain gap and avoid distorting intrinsic location information on decoder. We calculate the ratio between highly accurate boxes and accurate boxes by measuring the mAP ratio between a more strict (e.g., mAP60) and the default mAP50 criteria. The differences of this mAP ratio between ours and competitors are in Fig. \ref{fig:fig_mislocation}. Ours yields more highly accurate predictions than others.

\textbf{Detection results}. We illustrate some detection results of source model, SFA \cite{wang2021exploring} and ours, and the ground-truth boxes in Fig. \ref{fig:fig_qualitative}. Ours significantly reduce the false negatives, validating that O$^2$net can effectively diminish the domain gap. In addition, our approach shows advanced performance on small objects. This manifests that aligning multi-scale backbone features is more effective for detecting the small object than aligning the encoder embeddings as the CNN backbone features contains more local details for small objects.

\textbf{Feature Visualization}. Utilizing t-SNE \cite{tsne}, we visualize the feature distributions of CNN and decoder from two domains, as shown in Fig. \ref{fig:fig_tsne}. The features of the source model are separated by domains, while our method aligns the features of two domains on both CNN backbone and decoder, thus achieving performance gains. In addition, the decoder features have large intra-class variance since they carry location information. On the other hand, adversarial decoder alignment compresses the feature distribution of decoder, which might damage the intrinsic location information.

\section{Conclusion}
In this work, we delve into the problem of adapting the DETR-style detector to new domains. We build our method based on two findings, introducing an Object-Aware Alignment module for aligning the multi-scale backbone features and an Optimal Transport based Alignment module for reducing the domain gap on decoder features and maximumly preserving the localization information. Both modules contribute to the promising performance of our method. \vspace{-4mm}

\section*{Acknowledgements}
This paper was supported by National Key R\&D Program of China (No. 2021YFB3301503), and also supported by the National Natural Science Foundation of China under Grant No. U21A20519.


\bibliographystyle{ACM-Reference-Format}
\balance
\bibliography{reference}

\end{document}